\documentclass[]{template}

\usepackage{mathtools}
\usepackage{nicefrac}
\usepackage{float}
\usepackage{multirow}
\usepackage{mathrsfs}
\usepackage{manyfoot}
\usepackage{framed}
\usepackage{adjustbox}
\usepackage{subcaption}
\usepackage{cuted}
\usepackage{xspace}
\usepackage{makecell}
\usepackage{array}
\usepackage{ragged2e}
\newcolumntype{L}{>{\RaggedRight\hangafter=1\hangindent=0em}X}
\usepackage[capitalize,noabbrev]{cleveref}

\usepackage{xspace}
\usepackage{pgf}
\usepackage{colortbl}
\definecolor{softblue}{RGB}{90, 140, 200}
\providecommand{\heatb}[3]{}
\renewcommand{\heatb}[3]{%
  \pgfmathsetmacro{\hmdenom}{#3-#2}%
  \pgfmathparse{abs(\hmdenom)<0.00001}%
  \ifnum\pgfmathresult=1\relax
    \def\hmratio{0}%
  \else
    \pgfmathsetmacro{\hmratio}{min(1,max(0,(#1-#2)/\hmdenom))}%
  \fi
  \pgfmathtruncatemacro{\hmshade}{round(10 + 35*\hmratio)}%
  \edef\hmcolor{softblue!\hmshade!white}%
  \expandafter\cellcolor\expandafter{\hmcolor}#1%
}

\providecommand{\heatbd}[3]{}
\renewcommand{\heatbd}[3]{%
  \pgfmathsetmacro{\hmdenom}{#3-#2}%
  \pgfmathparse{abs(\hmdenom)<0.00001}%
  \ifnum\pgfmathresult=1\relax
    \def\hmratio{0}%
  \else
    \pgfmathsetmacro{\hmratio}{min(1,max(0,(#3-#1)/\hmdenom))}%
  \fi
  \pgfmathtruncatemacro{\hmshade}{round(10 + 35*\hmratio)}%
  \edef\hmcolor{softblue!\hmshade!white}%
  \expandafter\cellcolor\expandafter{\hmcolor}#1%
}

\providecommand{\gooddelta}[1]{\textcolor{green!45!black}{$#1$}}

\newcommand{\methodname}{\textsc{VP}\xspace}
\newcommand{\papertitle}{Scalable Visual Pretraining for Language Intelligence}

\newcommand{\ie}{\textit{i.e.}}
\newcommand{\eg}{\textit{e.g.}}

\definecolor{Gray}{gray}{0.9}

\makeatletter
\newcommand{\vpabstracttext}{}
\let\vp@templateabstract\abstract
\long\def\abstract#1{\gdef\vpabstracttext{\ignorespaces#1}}
\makeatother
\abstract{
The rapid progress of large foundation models has been driven predominantly by pretraining on large-scale text corpora.
However, many forms of knowledge are conveyed through visual representations, 
where figures, typeset equations, and page layouts carry rich information that cannot be faithfully or completely captured by text alone. 
Yet current pretraining approaches discard these visual cues by converting visually rich sources, 
such as documents and web pages, into plain text for learning language intelligence.
This paper challenges the default assumption that language models must be trained on text-only representations and shows that Visual Pretraining is a scalable learner for foundation model intelligence.
To this end, we conduct a systematic study of unsupervised visual pretraining paradigms that directly leverage visual documents without text extraction.
Across multiple backbones and benchmarks, 
visual pretraining on the same underlying corpora consistently outperforms text-only pretraining, 
offering an efficient pathway to scalable language intelligence.

}
\makeatletter
\let\abstract\vp@templateabstract
\makeatother

\title{\papertitle}

\author{{\normalfont\normalsize\bfseries Yiming Zhang$^{1,2,*}$, Zhonghan Zhao$^{1,3,*}$, Wenwei Zhang$^{1,*}$, Haiteng Zhao$^{1}$, Tianyang Lin$^{1}$,}\\
{\normalfont\normalsize\bfseries  Huanze Tang$^{1}$, Yunhua Zhou$^{1}$, Demin Song$^{1}$, Kuikun Liu$^{1}$, Haochen Ye$^{1}$, Haian Huang$^{1}$, }\\
{\normalfont\normalsize\bfseries Yuzhe Gu$^{1,4}$, Haijun Lv$^{1}$, Qipeng Guo$^{1}$, Bin Liu$^{2}$, Gaoang Wang$^{3,\dag}$, Kai Chen$^{1,\dag}$}\\
{\normalfont\normalsize $^1$Shanghai Artificial Intelligence Laboratory \quad
$^2$University of Science and Technology of China}\\
{\normalfont\normalsize $^3$Zhejiang University \quad
$^4$Shanghai Jiao Tong University}\\
{\normalfont\normalsize $^*$These authors contributed equally to this work. \quad
$^\dag$ Corresponding authors.}\\
{\normalfont\normalsize \texttt{gaoangwang@zju.edu.cn, chenkai@pjlab.org.cn}}
}

\begin{abstract}
\vpabstracttext
\end{abstract}

\begin{document}
\maketitle

\begin{figure*}[t]
  \centering
  \includegraphics[width=\linewidth]{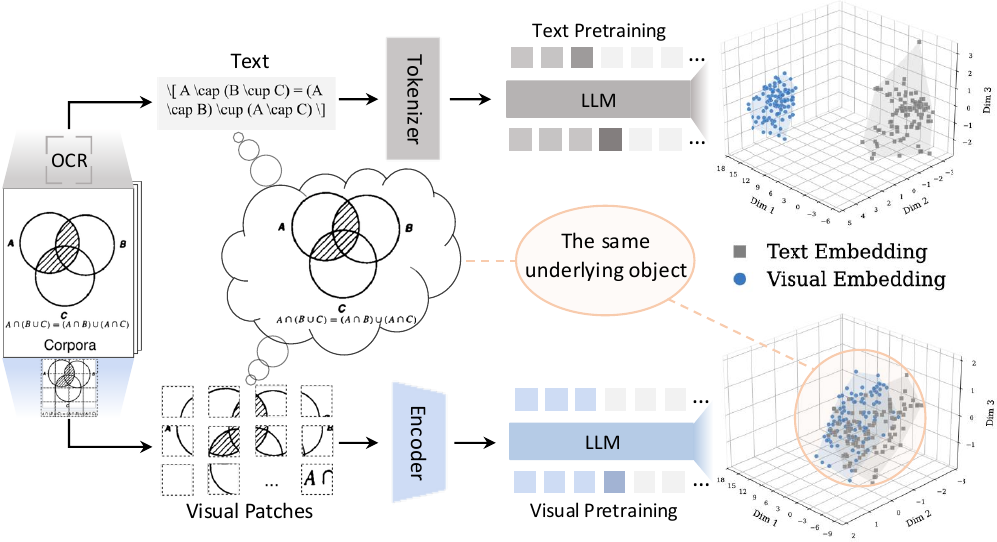} 
  \caption{\textbf{Matched text and visual pretraining from the same scientific-document corpus.}
  In the TP pathway, PDF pages are parsed into text, tokenized, and trained with text-token prediction, which can discard or distort page-level structure such as diagrams, equations, tables, and layout.
  In the VP pathway, the same pages are rendered as images, filtered into foreground visual patches, and trained with next visual latent prediction in a frozen visual-feature space.
  The t-SNE \citep{vandermaaten08a} projections on the right illustrate the representation change quantified in~\Cref{tab:alignment_multimodal}: VP brings matched visual and textual document embeddings closer in the shared representation space.}
  \label{fig:main}
\end{figure*}

\section{Introduction}
The striking advances of large foundation models have been driven by pretraining on 
text corpora at unprecedented scale \citep{brown2020language, kaplan2020scaling, hoffmann2022training}.
While effective, this paradigm rests on a strong implicit assumption: 
that all knowledge worth learning can be losslessly encoded as a linear sequence of text tokens.
However, cognitive science has long shown that humans routinely reason with
diagrams, spatial layouts, mathematical notation, and other representational
forms whose visual cues make certain relations directly available for
inference \citep{larkin1987diagram,zhang1997external,barsalou1999perceptual,barsalou2008grounded}.
Converting these visual cues into plain text before training is therefore inherently lossy.
Here we challenge this assumption and show that foundation models can 
learn directly from visual corpora without text extraction or image-text pairing supervision, 
yielding stronger language intelligence than text pretraining on the same underlying corpus.

Scientific documents are a particularly acute case of this loss.
Papers, textbooks, and technical reports communicate complex content through figures,
tables, formula layouts, and page-level spatial organization, all of which encode
geometric constraints, symbolic topologies, and structural correspondences essential
to scientific reasoning.
The recently proposed Platonic Representation Hypothesis \citep{huh2024platonic}
formalizes a related intuition for machine learning, arguing that representations learned
by different models and modalities converge toward shared abstractions of reality.
These observations imply that the linguistic information
extracted from a scientific document is only a projection of a richer underlying
structure, and that this structure could in principle be learned directly from raw
visual documents.

Yet existing approaches do not exploit this possibility along either of two dimensions.
At the data level, pretraining corpora reduce visual documents to plain text,
either by parsing HTML or LaTeX source \citep{paster2024openwebmath, gao2020pile}
or by applying neural document-parsing models to PDFs as a preprocessing step
\citep{blecher2024nougat, lee2023pix2struct, kim2021donut, wang2024mineru}, after
which the language model is trained exclusively on the resulting text.
At the training level, recent multimodal foundation models do incorporate visual modalities during training \citep{wang2024qwen2, chen2024internvl, liu2023visual},
but they treat visual inputs as conditioning context for text-token prediction, 
preventing visual content from being deeply integrated into the predictive process.
In both regimes the raw documents are consumed and then discarded leaving the visual modality outside the model's learning objective.

\begin{table*}[t]
\centering
\caption{\textbf{VP improves text-only scientific reasoning under matched document sources.}
All rows use the same starting checkpoint and SFT stage.
TP and VP differ only in how the scientific-PDF corpus is represented: MinerU2.5-parsed text for TP and rendered page images for VP.
Scores are reported on text-only reasoning benchmarks; GPQA denotes GPQA Diamond and AIME denotes AIME 2025.}
\label{tab:main_result_text}
\setlength{\tabcolsep}{5pt}
\renewcommand{\arraystretch}{1.08}
\resizebox{\linewidth}{!}{%
\begin{tabular}{lcccc c cccc}
\toprule
\multirow{2}{*}{\textbf{Method}}
& \multicolumn{4}{c}{\textbf{Multimodal Models}}
& \multicolumn{1}{c}{}
& \multicolumn{4}{c}{\textbf{Language Models}} \\
\cmidrule(lr){2-5}\cmidrule(lr){7-10}
& \textbf{MMLU-Pro} & \textbf{GPQA} & \textbf{AIME} & \textbf{HLE}
& 
& \textbf{MMLU-Pro} & \textbf{GPQA} & \textbf{AIME} & \textbf{HLE} \\
\midrule

& \multicolumn{4}{c}{\textit{Qwen 3.5} \citep{qwen3.5}}
&
& \multicolumn{4}{c}{\textit{Qwen 3} \citep{yang2025qwen3}} \\
\cmidrule(lr){2-5}\cmidrule(lr){7-10}
Base                
& \heatb{82.31}{82.31}{85.09} 
& \heatb{77.84}{76.24}{79.29} 
& \heatb{81.56}{81.56}{90.21}  
& \heatb{14.39}{14.39}{16.67} 
& 
& \heatb{81.21}{81.21}{81.94}    
& \heatb{75.06}{74.94}{77.08}   
& \heatb{75.44}{74.99}{76.98}   
& \heatb{10.59}{10.59}{11.77} \\

Text Pretraining    
& \heatb{83.91}{82.31}{85.09} 
& \heatb{76.24}{76.24}{79.29} 
& \heatb{89.58}{81.56}{90.21} 
& \heatb{15.70}{14.39}{16.67} 
& 
& \heatb{81.52}{81.21}{81.94} 
& \heatb{74.94}{74.94}{77.08} 
& \heatb{74.99}{74.99}{76.98}  
& \heatb{11.39}{10.59}{11.77} \\

\methodname\ (Ours) 
& {\heatb{85.09}{82.31}{85.09}} 
& {\heatb{79.29}{76.24}{79.29}} 
& {\heatb{90.21}{81.56}{90.21}} 
& {\heatb{16.67}{14.39}{16.67}} 
& 
& {\heatb{81.94}{81.21}{81.94}} 
& {\heatb{77.08}{74.94}{77.08}} 
& {\heatb{76.98}{74.99}{76.98}} 
& {\heatb{11.77}{10.59}{11.77}} \\
\midrule

& \multicolumn{4}{c}{\textit{Llama 3.2 Vision} \citep{meta2024llama32vision}}
&
& \multicolumn{4}{c}{\textit{Llama 3.1} \citep{grattafiori2024llama}} \\
\cmidrule(lr){2-5}\cmidrule(lr){7-10}
Base                
& \heatb{47.24}{47.24}{51.52} 
& \heatb{27.46}{27.46}{33.08} 
& \heatb{8.02}{8.02}{18.54} 
& \heatb{6.41}{6.08}{7.00} 
& 
& \heatb{59.48}{59.48}{62.77}    
& \heatb{39.71}{39.71}{47.10}  
& \heatb{24.17}{23.65}{24.27} 
& \heatb{6.75}{6.75}{7.17} \\

Text Pretraining    
& \heatb{50.60}{47.24}{51.52} 
& \heatb{30.24}{27.46}{33.08} 
& \heatb{13.75}{8.02}{18.54} 
& \heatb{6.08}{6.08}{7.00} 
& 
& \heatb{60.64}{59.48}{62.77} 
& \heatb{43.88}{39.71}{47.10} 
& \heatb{23.65}{23.65}{24.27} 
& \heatb{6.79}{6.75}{7.17} \\

\methodname\ (Ours) 
& {\heatb{51.52}{47.24}{51.52}} 
& {\heatb{33.08}{27.46}{33.08}} 
& {\heatb{18.54}{8.02}{18.54}} 
& {\heatb{7.00}{6.08}{7.00}} 
& 
& {\heatb{62.77}{59.48}{62.77}} 
& {\heatb{47.10}{39.71}{47.10}} 
& {\heatb{24.27}{23.65}{24.27}} 
& {\heatb{7.17}{6.75}{7.17}} \\
\bottomrule
\end{tabular}%
}
\end{table*}

Here we present Visual Pretraining (VP), a framework in which a
foundation model learns visual information directly from raw documents without any text extraction or image-text pairing supervision.
Through a systematic study under carefully matched data sources, we
find that across multiple LLM backbones and scientific reasoning benchmarks, VP
consistently outperforms text-only pretraining on the same underlying corpus, is
substantially more efficient (using only 25\% of the token budget) with respect to both model size and data scale, 
and strengthens cross-modal alignment, establishing VP as a scalable pathway for
learning both language and visual intelligence.
This work makes three contributions. 
First, we show that reducing visual documents to plain text incurs substantial information loss, 
and that visual pretraining recovers this otherwise discarded information.
Second, we introduce an autoregressive visual pretraining framework that trains a foundation model to predict document patches in latent space, 
deeply integrating visual latent into the predictive process. 
Third, through a unified empirical study with matched corpora across multiple architectures and
benchmarks, we establish VP as an effective, efficient, and scalable alternative to text-only pretraining.

\section{Results}
\label{sec:results}

\begin{figure*}[t]
  \centering
  \includegraphics[width=\linewidth]{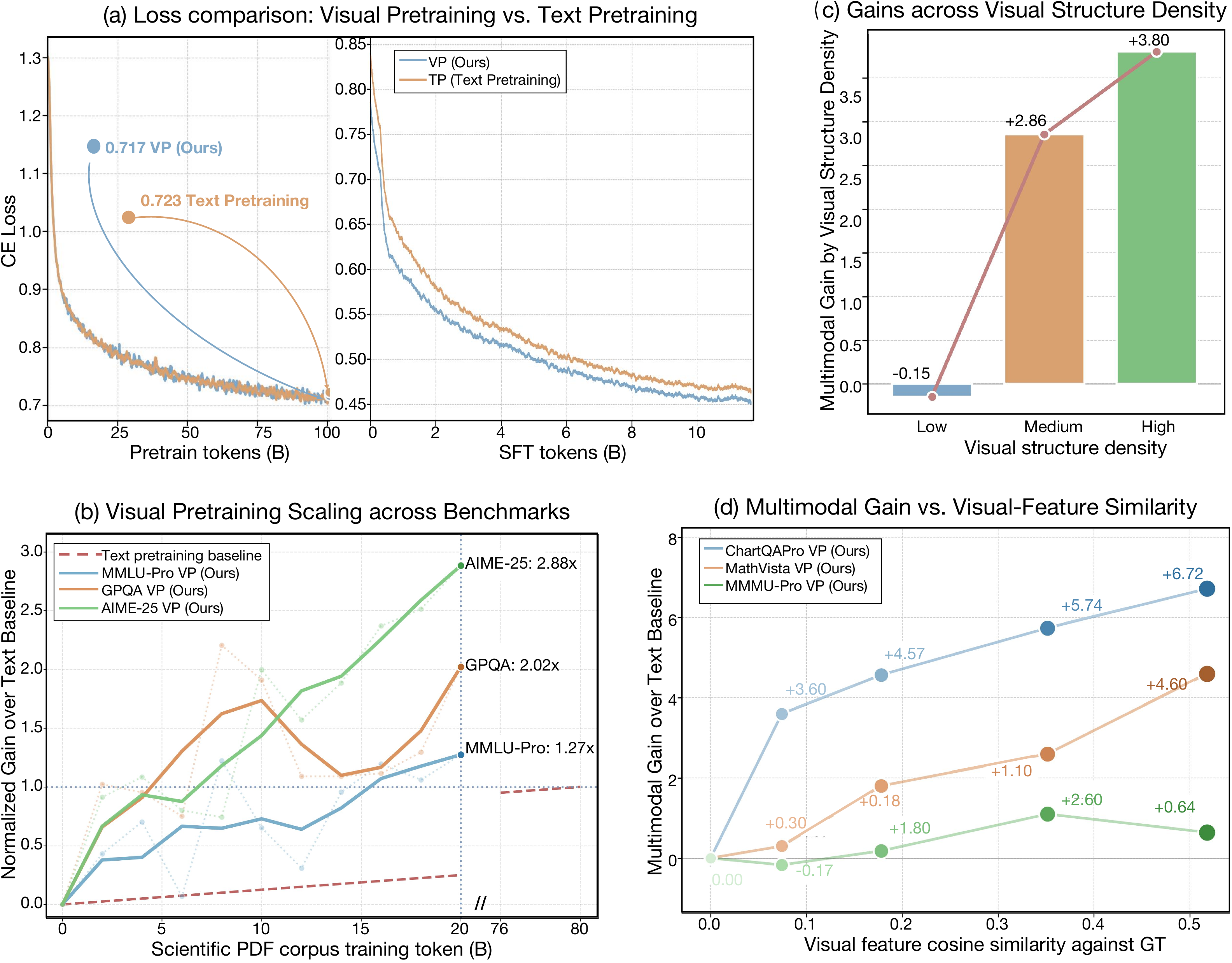}
  \caption{
  \textbf{Visual pretraining scales with retained PDF visual tokens and benefits most from structure-heavy pages.}
  \textbf{(a)}, VP reaches a more favorable SFT trajectory after comparable CPT loss.
  \textbf{(b)}, Increasing the retained visual-token budget consistently raises downstream gains, normalized by the TP-over-base improvement.
  \textbf{(c)}, VP's advantage is largest on examples with high visual-structure density, where figures, equations, tables, and layout carry more of the evidence.
  \textbf{(d)}, Better next-token visual-feature prediction is associated with larger multimodal gains over TP, linking visual-space modeling quality to downstream transfer.
  }
  \label{fig:scaling}
\end{figure*}
Our experiments establish three findings. 
First, visual pretraining (VP) outperforms text pretraining (TP) on scientific reasoning under matched corpora.
% Second, this gain scales efficiently with training data. 
% Second, on the same underlying corpus, VP outperforms TP with only 25\% of the token budget.
Second, this gain scales efficiently with training data. 
On the same underlying document corpus, VP surpasses TP while consuming only 25\% of the token budget.
Third, without any image–text pair supervision, VP improves visual perception as well as reasoning across modalities.
Throughout the experiments, 
VP and TP use the same text corpus and a matched additional scientific-PDF corpus in the continued pretraining (CPT) stage.
VP trains on raw document pages represented as visual tokens, 
whereas TP trains on parsed text from the same documents. 
This setup isolates the effect of preserving the native visual form of scientific documents.
Then we apply identical supervised fine-tuning (SFT) to all VP and TP checkpoints before evaluation 
to elicit the structured reasoning traces and answer formats expected by the benchmarks.

\paragraph{Effectiveness: \methodname improves scientific language reasoning}
\label{sec:results_main}

We first establish that continued pretraining on raw document pages yields stronger performance on scientific
reasoning benchmarks than continued pretraining on textualized documents from the same corpus. 
We compare the base model, the matched TP baseline, and \methodname under identical document sources and SFT data. 
The two pretraining settings differ only in how documents are represented.
Experiments are conducted across state-of-the-art foundation models, including Qwen3.5 \citep{qwen3.5}, Qwen3 \citep{yang2025qwen3}, Llama3.2 Vision \citep{meta2024llama32vision} and Llama3.1 \citep{grattafiori2024llama}.
% TP is trained on the text extracted from each page, whereas \methodname is trained on the native page, 
% including its layout, equations, tables, figures and other visually organized structures. 
Unless otherwise specified, we report average pass@8 for GPQA \citep{rein2023gpqa}, 
the average score over 32 runs for AIME-25 \citep{AIME2025}, 
and pass@1 for MMLU-Pro \citep{wang2024mmlu} and HLE \citep{phan2025humanity}. 
Results are reported in~\Cref{tab:main_result_text}.

% We first show that continued pretraining on raw document pages yields stronger performance on scientific
% reasoning benchmarks than continued pretraining on textualized documents from the same corpus.
% As shown in~\Cref{tab:main_result_text}, 
% \methodname consistently improves over matched text pretraining baseline. 
% Unless otherwise specified, we report average pass@8 for GPQA, the average score over 32 runs for AIME-25, and pass@1 for MMLU-Pro and HLE. 
% The gains appear across model families, including Qwen and Llama variants, indicating that the benefit is not tied to a single architecture.

\methodname consistently improves over the matched TP baseline, and the gains appear across
both native multimodal models and language-only models, indicating that the benefit is not tied to a single architecture. 
Because the two settings differ only in document representation, the improvement on reasoning benchmarks
suggests that VP is an effective way to acquire language intelligence directly from visual corpora, which
preserve higher-fidelity reasoning-relevant information that text extraction weakens or discards.
Specifically, the consistent improvements on multi-disciplinary benchmarks indicate that \methodname
successfully internalizes knowledge from the visual content of scientific documents. 
GPQA Diamond improves by up to 3.22 points across the four backbones (e.g., 76.24 to 79.29 on Qwen 3.5) and MMLU-Pro by up to 2.1 points, 
consistent with the view that source-document knowledge in these scientific domains is partly carried by geometric
figures, physics schematics and other visual content that text extraction cannot faithfully preserve.
HLE, by contrast, shows only marginal improvements (up to 0.97 points), 
since its difficulty derives primarily from hard multi-step reasoning
rather than the knowledge gained from visual information. 
\begin{figure*}[t]
  \centering
  \includegraphics[width=\linewidth]{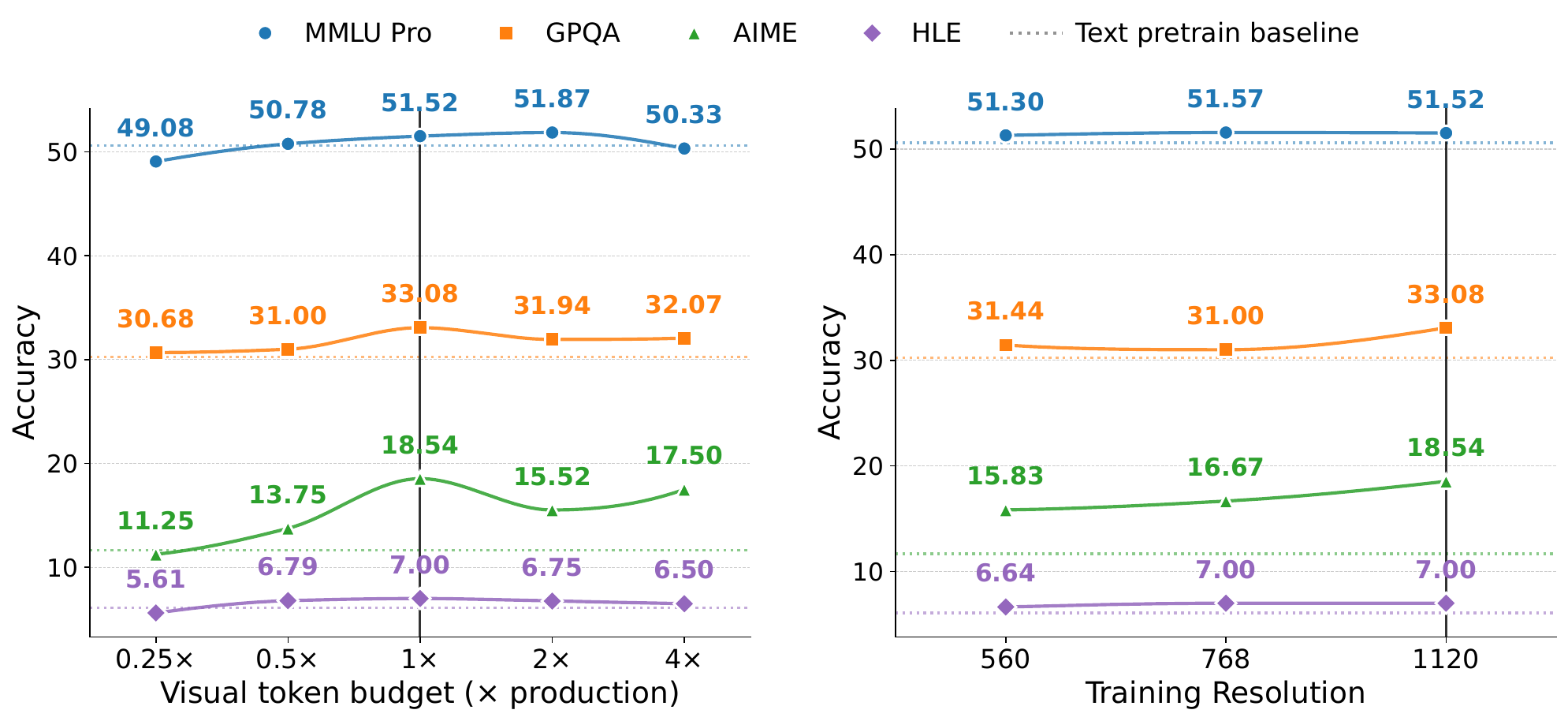}
  \caption{
  \textbf{The main VP setting balances visual-token budget and rendering resolution.}
  % \textbf{Left}, within a fixed training context, varying the visual-token budget shows that a 1$\times$ setting (8,192 foreground tokens) achieves the best trade-off. 
  \textbf{Left}, with the training context fixed for each forward pass, varying the visual token budget shows that the 1$\times$ setting (8,192 foreground visual tokens per batch) offers the best trade-off.
  % Beyond this point, longer visual sequences within each batch inflate the proportion of spatially adjacent negatives in the optimization objective, producing suboptimal returns.
  % Beyond this point, longer visual sequences within each batch inflate hard-negative saturation in the optimization objective, raising gradient variance and producing suboptimal returns.
  % \textbf{Left}, within a fixed training context, varying the visual--text token allocation shows that a 1$\times$ visual budget (8,192 foreground tokens) achieves the best trade-off.
  % Larger visual budgets displace text tokens and become suboptimal.
  \textbf{Right}, with the visual budget fixed at 8,192 tokens, 
  lowering the maximum resolution preserves strong performance, 
  indicating the efficiency of VP under compressed visual input.
  % \textbf{Left}, at 1120 resolution, the visual--text token ratio is varied under a fixed context budget. The 1$\times$ visual setting (8,192 tokens) is optimal. Beyond this point, gains vanish as text tokens are displaced.
  % \textbf{Right}, With the visual context fixed at 8,192 tokens, lowering the rendering resolution preserves strong performance, indicating that VP does not rely solely on exhaustive high-resolution rendering.
  }
  \label{fig:ablation_ratio}
\end{figure*}
\paragraph{Scalability: gains scale with \methodname }
\label{sec:results_scaling}

We further demonstrate that the benefit of \methodname is efficient and scalable. 
\Cref{fig:scaling}(a) compares the training loss dynamics of \methodname and TP across CPT and SFT stages.
% During CPT, the two curves are close, with \methodname reaching a slightly lower final loss. 
% After both checkpoints are trained with the same SFT data, 
% the \methodname-initialized model converges faster and reaches a lower SFT loss. 
During CPT, the two curves are close, with \methodname reaching a slightly lower final loss. 
After both checkpoints are fine-tuned with the same SFT data, 
the \methodname checkpoint converges faster and reaches a lower final SFT loss.
% We attribute this phenomenon to that measure different aspects of the model.
% We attribute this phenomenon to the fact that loss in pretraining stage does not adequately capture downstream capabilities~\citep{wei2022emergent}. 
% SFT loss, by contrast, is contingent upon the geometric organization established during pretraining~\citep{li2025tracing}, 
% and therefore more directly reflects the quality of the learned representation and its readiness for downstream reasoning.
We attribute this discrepancy to the limited sensitivity of pretraining loss to downstream capabilities~\citep{wei2022emergent}. 
SFT loss, by contrast, is grounded in the geometric structure established during pretraining~\citep{li2026tracing}, 
and thus more directly reflects representation quality and its capacity for reasoning.

We then examine how downstream performance scales with training dynamics. 
% \Cref{fig:scaling}(b) shows that \methodname achieves steadily increasing normalized gains as training progresses, outperforming text pretraining in scaling efficiency. 
In \Cref{fig:scaling}(b), we normalize the downstream gain of VP against that of TP, both measured relative to the starting checkpoint. 
For the same PDF corpus, VP trains on only approximately 20B visual tokens, while TP processes roughly 80B text tokens.
Moreover, VP achieves steadily increasing normalized gains as training progresses, 
reaching 1.27× on MMLU-Pro, 2.02× on GPQA and 2.88× on AIME-25, indicating superior scaling efficiency.
\Cref{fig:scaling}(d) further reveals that visual feature cosine similarity, 
a proxy for visual prediction quality, correlates strongly with downstream improvements, 
suggesting that better visual space modeling supports stronger reasoning, by enabling richer knowledge extraction from visual content.
To verify that these gains are visually grounded, 
we categorize the evaluation sets by visual-structure density following \citet{masry2025chartqapro} (\Cref{fig:scaling}(c)). 
The advantage of VP over TP grows with density. 
On low-density, text-dominant pages, VP and TP perform on par, whereas high-density pages rich in figures,
equations, and tables produce substantially larger improvements. 
This confirms that VP's scaling benefits are genuinely driven by its modeling of visual content.

\paragraph{Efficiency: compact visual representations preserve VP gains}

We next demonstrate the efficiency of VP.
First, visual representations compress the same scientific documents far more effectively than OCR-extracted text.
Specifically, VP trains on approximately 20B visual tokens, whereas TP consumes roughly 80B parsed text tokens.
Second, under a fixed optimization token budget, VP achieves strong gains with only a modest share of visual tokens.
% We vary the visual-token budget while fixing the overall training mixture (\Cref{fig:ablation_ratio}, left) and adopt a text-to-visual ratio of 4:1 (1$\times$) as our main setting.
% At this ratio, visual tokens account for only a modest fraction of the training context, yet this small share suffices to deliver gains.
% When the visual-token ratio exceeds 4:1, performance becomes suboptimal. 
% We attribute this saturation to the model shifting capacity toward visually dense regions and away from text, consistent with the density-stratified results in \Cref{fig:scaling} (c).
% The 8,192-token setting (1$\times$) achieves the best trade-off, yet extending the visual sequence beyond this length produces diminishing and eventually suboptimal performance. 
% We attribute the suboptimal performance at larger budgets to the changed optimization dynamics of the visual InfoNCE objective, 
% as increasing the number of visual tokens also changes the number of contrastive terms and effective negatives. 
We vary the visual token budget within a fixed training context (\Cref{fig:ablation_ratio}, left). 
The 8,192-token setting (1$\times$) achieves the best trade-off.
We hypothesize that the suboptimal performance at larger visual-token budgets stems from optimization effects in the visual InfoNCE training. 
When more visual tokens are retained, the loss is computed over more prediction targets and in-batch contrastive terms, which can change the effective gradient scale and optimization behavior.
Moreover, because we keep the learning rate and other optimization hyperparameters fixed in this sweep, larger token budgets may not be fully tuned under this training recipe.
% We attribute this to negative-sample saturation in the visual InfoNCE objective. 
% Longer visual sequences inflate the proportion of hard, intra-page negatives, amplifying gradient variance and degrading optimization stability.
Finally, the visual input itself admits further compression. 
We train VP at varying resolutions and evaluate downstream performance.
With the visual budget fixed at 8,192 tokens, 
lowering the maximum resolution preserves strong performance (\Cref{fig:ablation_ratio}, right), and VP still outperforms TP.
This further amplifies the efficiency of VP.

\paragraph{Cross-modality: multimodal capabilities emerge without paired supervision}
\label{sec:results_alignment}
We finally demonstrate that the gains from \methodname generalize beyond text-only reasoning. 
Although \methodname is performed on unlabeled raw document pages, without task-specific multimodal annotations,
it also improves multimodal representation and downstream visual reasoning.

\begin{table*}[t]
\centering
\caption{\textbf{VP improves cross-modal alignment and multimodal transfer without labeled multimodal pretraining data.}
Left \textbf{(a)}: alignment metrics on 100 held-out scientific document image--text pairs before and after VP; lower centroid separation and higher cosine similarity, CKA, and mutual $k$-NN indicate better alignment.
Right \textbf{(b)}: pass@1 performance on multimodal benchmarks for native multimodal models, comparing the base checkpoint, TP, and VP.
The gains show that unlabeled visual-document pretraining improves both representation compatibility and downstream visual reasoning.}
\label{tab:alignment_multimodal}
\small
\setlength{\tabcolsep}{5pt}
\renewcommand{\arraystretch}{1.08}

\begin{minipage}[t]{0.383\linewidth}
\centering
\subcaption{\textbf{Cross-modal alignment.}}
\label{tab:alignment}
\vspace{0pt}
\resizebox{\linewidth}{!}{%
\begin{tabular}{@{}lccc@{}}
\toprule
\textbf{Metric} 
& \textbf{Original} 
& \textbf{\methodname\ (Ours)} 
& $\Delta$ \\
\midrule

\multicolumn{1}{c}{} 
& \multicolumn{3}{c}{\textit{Global}} \\
\cmidrule(lr){2-4}
Centroid Sep.\,$\downarrow$ 
& \heatbd{1.665}{0.661}{1.665} 
& {\heatbd{0.661}{0.661}{1.665}} 
& \gooddelta{-1.004} \\

Cosine Sim.\,$\uparrow$     
& \heatb{0.631}{0.631}{0.907}     
& {\heatb{0.907}{0.631}{0.907}}     
& \gooddelta{+0.276} \\

\midrule
\multicolumn{1}{c}{} 
& \multicolumn{3}{c}{\textit{Structural}} \\
\cmidrule(lr){2-4}
Linear CKA\,$\uparrow$      
& \heatb{0.657}{0.657}{0.745}      
& {\heatb{0.745}{0.657}{0.745}}      
& \gooddelta{+0.088} \\

\midrule
\multicolumn{1}{c}{} 
& \multicolumn{3}{c}{\textit{Local}} \\
\cmidrule(lr){2-4}
Mutual $k$-NN@1\,$\uparrow$  
& \heatb{0.140}{0.140}{0.310}  
& {\heatb{0.310}{0.140}{0.310}}  
& \gooddelta{+0.170} \\

Mutual $k$-NN@5\,$\uparrow$  
& \heatb{0.288}{0.288}{0.420}  
& {\heatb{0.420}{0.288}{0.420}}  
& \gooddelta{+0.132} \\

Mutual $k$-NN@10\,$\uparrow$ 
& \heatb{0.395}{0.395}{0.496} 
& {\heatb{0.496}{0.395}{0.496}} 
& \gooddelta{+0.101} \\

\bottomrule
\end{tabular}%
}
\end{minipage}
\hfill
\begin{minipage}[t]{0.595\linewidth}
\centering
\subcaption{\textbf{Multimodal benchmark performance.}}
\label{tab:main_result_multimodal}
\vspace{0pt}
\resizebox{\linewidth}{!}{%
\begin{tabular}{@{}lcccc@{}}
\toprule
\textbf{Method} 
& \textbf{MMMU-Pro} 
& \textbf{SFE} 
& \textbf{ChartQAPro} 
& \textbf{MathVista} \\
\midrule

\multicolumn{1}{c}{} 
& \multicolumn{4}{c}{\textit{Qwen 3.5} \citep{qwen3.5}} \\
\cmidrule(lr){2-5}
Base                
& \heatb{71.39}{71.39}{73.87}    
& \heatb{53.41}{53.09}{56.57}    
& \heatb{57.99}{56.42}{61.80}    
& \heatb{84.30}{84.30}{86.70} \\

Text Pretraining    
& \heatb{72.14}{71.39}{73.87} 
& \heatb{53.09}{53.09}{56.57} 
& \heatb{56.42}{56.42}{61.80} 
& \heatb{85.50}{84.30}{86.70} \\

\methodname\ (Ours) 
& {\heatb{73.87}{71.39}{73.87}} 
& {\heatb{56.57}{53.09}{56.57}} 
& {\heatb{61.80}{56.42}{61.80}} 
& {\heatb{86.70}{84.30}{86.70}} \\

\midrule
\multicolumn{1}{c}{} 
& \multicolumn{4}{c}{\textit{Llama 3.2 Vision} \citep{meta2024llama32vision}} \\
\cmidrule(lr){2-5}
Base                
& \heatb{28.55}{28.21}{29.19} 
& {\heatb{28.86}{28.86}{31.08}} 
& \heatb{20.95}{20.95}{27.67} 
& {\heatb{39.80}{39.60}{44.40}}  \\

Text Pretraining    
& \heatb{28.21}{28.21}{29.19} 
& {\heatb{29.36}{28.86}{31.08}} 
& \heatb{22.23}{20.95}{27.67} 
& {\heatb{39.60}{39.60}{44.40}}  \\

\methodname\ (Ours) 
& \heatb{29.19}{28.21}{29.19} 
& {\heatb{31.08}{28.86}{31.08}} 
& {\heatb{27.67}{20.95}{27.67}} 
& {\heatb{44.40}{39.60}{44.40}}  \\

\bottomrule
\end{tabular}%
}
\end{minipage}
\end{table*}

We extract hidden states from the foundation model (\ie, Qwen3.5) on 100 image–text pairs sourced from scientific documents, 
and compare the pooled visual and text embeddings before and after \methodname. 
\Cref{tab:alignment_multimodal} (a) shows consistent gains across three complementary levels of alignment. 
Globally, the centroid separation \citep{liang2022mind} drops from 1.665 to 0.661 and the paired cosine similarity rises from 0.631 to 0.907, 
indicating that the two modalities now share a common subspace and that paired samples are systematically co-located. 
Structurally, Linear CKA \citep{kornblith2019similarity} improves from 0.657 to 0.745, 
evidencing agreement on the larger geometric structure of the two representation spaces. 
Locally, Mutual k-NN overlap \citep{huh2024platonic} improves at all evaluated k, confirming neighbourhood alignment at the instance level. 
We also provide a qualitative view in \Cref{fig:main}, 
where the originally disjoint visual and text embedding clusters collapse into a single overlapping region after \methodname pretraining.
For visual comparability and fairness, both panels share anchor features and t-SNE settings.

% ============TODO============
\Cref{tab:alignment_multimodal} (b) reports multimodal benchmark (\ie, MMMU-Pro \citep{yue2025mmmu}, SFE \citep{zhou2026scientists}, ChartQAPro \citep{masry2025chartqapro} and MathVista \citep{lu2024mathvista}) performance with Qwen 3.5 and Llama 3.2 Vision. 
% \methodname consistently outperforms both the pretrained base and the baseline after text pretraining on every benchmark and backbone.
\methodname consistently outperforms both the pretrained base and the text-pretrained model on every benchmark and backbone.
TP yields negligible gains and occasionally regresses (\eg, $57.99 \to 56.42$ on ChartQAPro for Qwen 3.5), indicating that   
further TP cannot harvest information from visually rich documents. The VP advantage over TP is largest on
visually heavy benchmarks, reaching $+5.4$ on ChartQAPro for both backbones and $+4.8$ on MathVista for Llama 3.2 Vision.
Together with the density-stratified results in \Cref{fig:scaling} (c), this indicates that \methodname{}'s main benefit lies in
visual content (figures, equations, tables, and complex layouts) that text parsing cannot faithfully recover.

\begin{figure*}[t]
  \centering
  \includegraphics[width=\textwidth]{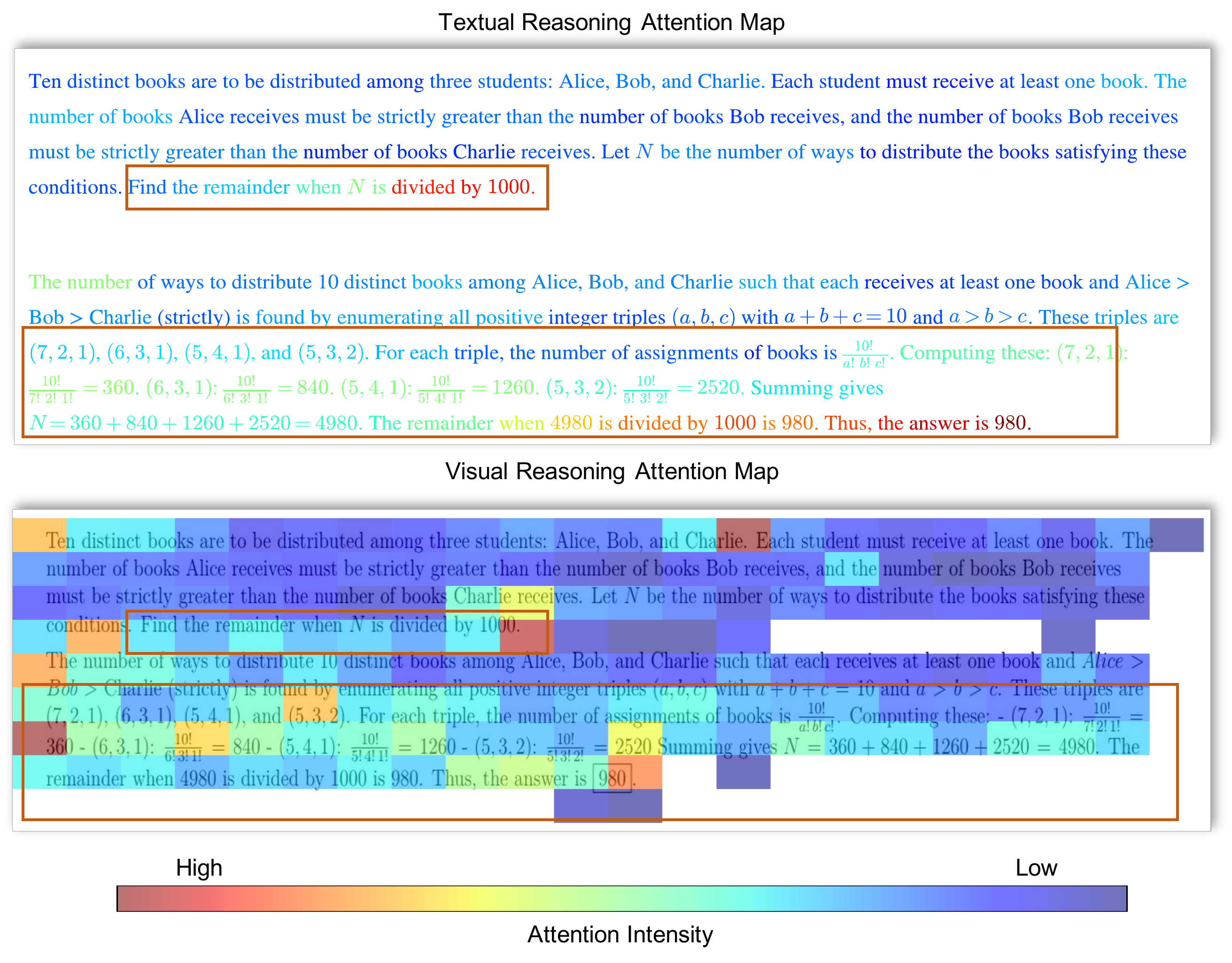}
  \caption{\textbf{Visual-token reasoning attends to the same semantic evidence as text reasoning.}
  We compare foundation models with TP or VP on the same math problem under text and image inputs.
  \textbf{Top}, textual reasoning: the problem and chain-of-thought solution are given as plain text, and attention is measured from the final answer sentence to previous tokens.
  \textbf{Bottom}, visual reasoning: the same problem is rendered as an image and fed through the visual pathway, and attention is measured from visual tokens in the answer region to retained page patches.
  Both views concentrate on the semantic regions highlighted in red, including the question constraint and intermediate computation steps.
  }
  \label{fig:attn_map}
\end{figure*}
\paragraph{Reasoning cues in visual pretraining}

To understand the mechanism behind VP's scientific reasoning gains, we conduct a qualitative study of attention patterns.
We compare attention maps elicited by the same math problem under text and visual inputs, using Qwen3.5 as the native multimodal backbone.
% As a qualitative probe, we compare attention maps elicited by the same math problem under text and visual inputs, using Qwen3.5 as the native multimodal backbone.
In the textual setting, the problem and its chain-of-thought solution are provided as plain text; we extract attention from the last Transformer layer, average over heads, and use the final answer sentence as the query.
% In the visual setting, the same problem is rendered as a image and fed through the visual pathway; we use visual tokens in the answer region as queries and average their attention across retained patches.
In the visual setting, the same problem is rendered as an image and fed through the visual pathway. We use visual tokens in the answer region as queries and average their attention across retained patches.
% We show a representative example in which both attention maps concentrate on corresponding semantic regions, 
We show a representative example where both attention maps highlight corresponding semantic regions,
including the question constraint and intermediate computation steps, while de-emphasizing filler content (\Cref{fig:attn_map}).
This probe does not imply that the two modalities use identical internal mechanisms, 
but it shows that VP exposes reasoning-relevant visual evidence to the shared LLM in a way that parallels the evidence used in text-based reasoning.

% \paragraph{Ablation study}

\section{Discussion}
\label{sec:discussion}
% \paragraph{Visual pretraining from native scientific documents}
Large language models typically acquire language intelligence from text-only corpora, 
yet much scientific knowledge is originally expressed in multimodal documents. 
Scientific papers contain equations, figures, tables and carefully designed layouts, 
all of which can be weakened or distorted when converted into plain text. 
We asked whether language models can acquire stronger scientific reasoning by directly adding visual pretraining on the native form of such documents to standard text continued pretraining.
% We asked whether language models can acquire stronger scientific reasoning by pretraining directly on the native visual form of such documents. 
% Our results provide a cautiously affirmative answer, showing that \methodname improves scientific reasoning over matched text pretraining, scales efficiently with compact visual budgets, and transfers to multimodal reasoning without labelled cross-modal data.
% The central finding is that visual pretraining can turn raw scientific pages into unlabeled supervision for pretraining foundation models toward stronger scientific reasoning and multimodal transfer.
Our results provide a cautiously affirmative answer, showing that \methodname improves scientific reasoning over matched text pretraining, 
scales efficiently with compact visual budgets, and transfers to multimodal reasoning without labeled cross-modal data. 
Together, these results show that raw scientific pages can provide unlabeled visual supervision that complements continued pretraining for foundation models.
The benefit of \methodname appears to come from preserving and modeling structure that is lost during textualization. 
Scientific pages are not arbitrary images, as their visual organization encodes equation topology, table structure, figure-text correspondence, symbolic continuity and spatial grouping.
\methodname trains the shared autoregressive backbone on foreground visual tokens ordered by their page positions, 
using a causal objective that predicts the next visual latent.
By making the visual document itself the prediction target, 
this objective encourages the backbone to model visual representations in a way that preserves document-native relations.
% \methodname places predictive pressure on these structures by requiring the shared autoregressive backbone to organize foreground document patches and predict future visual latents. 
% This objective differs from simply feeding images as auxiliary context, because the visual document itself becomes the target of pretraining.
Several observations support this interpretation. 
First, \methodname improves scientific reasoning over matched text pretraining across backbones, suggesting that the model can acquire reasoning-relevant knowledge from visual elements such as equations, figures, tables and layouts. 
Second, in multimodal evaluations, the gains are largest on visually dense examples, 
where figures, equations, tables and layout carry more of the evidence needed for reasoning.
Cross-modal alignment also improves without paired image-text supervision, 
suggesting that the visual objective reshapes the shared representation space in a way that becomes more compatible with language. 
Finally, the qualitative attention analysis suggests that visual-token reasoning can attend to semantically relevant regions of the rendered problem, 
although this should be read as supporting evidence rather than a proof that text and visual reasoning use identical mechanisms.

% \paragraph{Relation to existing pretraining methods}

Text continued pretraining remains an effective way to absorb knowledge that is already well represented in language, 
but scientific documents reach the model through a textualization pipeline that can discard layout, formulas, diagrams and other visual relations. 
\methodname does not replace this pathway. 
Instead, it mixes standard text continued pretraining with a visual next-latent objective on native scientific pages, 
allowing the model to retain text exposure while learning from the native visual form of the same scientific-document corpus.
Because \methodname and the text baseline use the same document source, 
the gain suggests that the native visual representation provides useful supervision beyond the textualized content alone.
% Text continued pretraining remains an effective way to absorb knowledge that is already well represented in language. 
% \methodname and the text baseline use the same PDF corpus, 
% so the gain reflects the native visual representation in which the content is learned rather than additional scientific content.

Existing multimodal pretraining methods usually introduce visual inputs through image-text pairs, captioning losses, contrastive alignment, OCR supervision or document-understanding objectives. OCR-free document models
such as Donut \citep{kim2021donut} and Pix2Struct \citep{lee2023pix2struct} further show that document images can be used directly, but their pretraining is mainly optimized for parsing, understanding or image-to-text generation. 
% In contrast, \methodname treats rendered scientific pages as unlabeled visual sequences and trains the shared autoregressive backbone with a next-latent prediction objective, without dense annotations or explicit image-text pairing.
In contrast, \methodname treats rendered scientific pages as unlabeled visual sequences and trains the shared autoregressive backbone with a next-latent prediction objective. 
It does so without dense annotations or explicit image-text pairing.

% \paragraph{Limitations}
Our study has several limitations. 
First, our approach should not be interpreted as a form of visual pretraining that is fully independent of language pretraining. 
Since \methodname is grounded in text pretraining and introduces visual-latent prediction during continued pretraining, the observed gains should be understood as evidence that visual pretraining can complement and extend foundation model pretraining.
Second, while \methodname shows that visual-latent prediction can effectively guide pretraining optimization, 
future work may further investigate how to better coordinate visual-latent decoding with text decoding. 
Our observations suggest that visual pretraining strongly encourages the model to capture visual presentations and structures, 
motivating future studies on loss scheduling, foreground-aware token selection, and parameter efficient visual modules.
Third, our experiments primarily focus on high knowledge-density scientific PDFs, where visual layouts, figures, tables, formulas, and surrounding textual context are tightly coupled to semantic meaning. 
% It remains an open question whether the same pretraining strategy yields intelligence on broader visual corpora such as natural images or videos data.
It remains an open question whether the same pretraining strategy transfers to broader visual corpora such as natural images or video.
This work points to two promising directions. 
First, visual pretraining provides a new route for foundation model training. 
For high-density visually native corpora, such as scientific documents, charts, tables, formulas, and structured layouts, visual-latent prediction may serve as an effective complementary pretraining objective. 
Second, visual pretraining opens a potential path toward a more scalable pretraining paradigm based on highly compressed visual corpora. 
This suggests that foundation models could be trained primarily on large-scale visual streams, with only lightweight text alignment, to match the performance of text-pretrained models.
% Future work will investigate whether this paradigm can enable efficient training over substantially larger data scales while retaining strong textual and multimodal capabilities.
Future work should investigate whether this paradigm enables efficient training at larger scales while preserving strong textual and multimodal capabilities.

\section{Methods}
\label{sec:methods}

\paragraph{Overview.}
% \methodname extends text continued pretraining with next visual latent prediction on raw scientific document pages. 
% Each page is encoded by a frozen vision tower; foreground features are filtered, kept in raster-scan order with spatial information preserved, projected into the LLM hidden space, and processed by the same autoregressive backbone used for text. 
% Under a causal image-only mask, the model predicts the next foreground feature in the frozen visual-feature space while continuing standard next-token text pretraining. 
% Because supervision is defined over frozen visual latents, \methodname requires neither OCR, image-text annotations, nor a pixel-level decoder; the matched TP baseline instead trains on MinerU2.5-parsed text from the same PDF pages~\citep{niu2025mineru25}. A full mathematical description of the pipeline and implementation details are provided in supplementary section~\Cref{sec:appendix_pipeline,sec:appendix_details}.
\methodname extends text continued pretraining by introducing next visual latent prediction on raw scientific document pages.
Each page is encoded by a frozen vision tower. 
Retained foreground features are then projected into the LLM's hidden space via a visual projection module.
We remove blank background regions, keep foreground patches, and order the remaining features in raster-scan order. 
Positional encodings are preserved to maintain spatial layout. 
The resulting sparse visual sequence is fed into the same autoregressive backbone that processes text tokens.
The model is trained to predict the next foreground patch feature under causal image-only masking.

\paragraph{Sparse document representation}
Given a rendered document page $\mathcal{I}$, the frozen vision tower $E_{\mathrm{v}}$ produces a sequence of visual features
\begin{equation}
\mathcal{Z}=E_{\mathrm{v}}(\mathcal{I})=(z_1,\ldots,z_N).
\label{eq:visual_features}
\end{equation}
Document pages contain large blank regions, such as margins and whitespace. We therefore compute a foreground mask using simple patch-level statistics, including pixel variance and average luminance, and retain only non-blank patches. The retained features are ordered in raster scan to form a sparse foreground sequence
\begin{equation}
\mathcal{U}
=
\operatorname{Raster}\{z_i:m_i=1\}
=
(u_1,\ldots,u_L), L\ll N .
\label{eq:foreground_sequence}
\end{equation}
This sparse representation keeps foreground document content in page order while substantially shortening the visual context. Each foreground feature is projected into the LLM hidden space through a learned linear projection. Position indices are re-assigned according to the raster order. The exact foreground filtering rule and sequence-packing mask are described in Supplementary~\Cref{sec:appendix_pipeline}.

\paragraph{Next visual latent prediction}
For each document image, we first extract a sequence of frozen visual latents and retain the foreground tokens.
These visual latents are mapped into the LLM embedding space by an input projection module and then fed to the LLM under a causal attention mask over visual positions.
Given the LLM hidden state at position $t$, an output projection head maps it back to the frozen visual-latent space to produce a prediction $\hat{\mathbf{z}}_{t+1}$.
The prediction is trained to match the next foreground visual latent $\mathbf{z}_{t+1}$ using a contrastive loss with in-batch negatives.
This forms an autoregressive visual-latent prediction objective. 
It follows the next-token prediction structure of language modeling, 
but the targets are continuous visual latents rather than discrete text tokens.
% The projected sequence is fed into the LLM with causal attention over image positions. 
% For each position, the hidden state is mapped by a lightweight prediction head to the frozen visual-feature target space. The predicted feature is trained to match the next foreground feature, making the visual task analogous to next-token prediction in language modelling, but over continuous document features.

We train the visual stream using a next-visual-latent prediction objective. For a batch of predicted--target pairs, let $p_{ij}$ be the softmax probability of matching the prediction at position $i$ to target feature $j$, computed from cosine similarities with temperature $\tau$. The visual pretraining loss is
\begin{equation}
\mathcal{L}_{\mathrm{VP}}
=
-\frac{1}{|\mathcal{B}|}
\sum_{i\in\mathcal{B}}
\log p_{ii}.
\label{eq:VP_loss}
\end{equation}
Other visual features in the batch serve as negatives. This objective encourages the model to predict the correct next document feature while distinguishing it from other patches. The expanded InfoNCE~\citep{oord2018representation} formulation and the definition of the in-batch matching probabilities are provided in Supplementary~\Cref{sec:appendix_pipeline}.

% We train the visual stream using a next-visual-latent prediction objective.
% For a set of valid predicted--target pairs $\mathcal{B}$, let
% $p_{ij}$ denote the probability of matching the prediction for pair $i$
% to the target visual latent of pair $j$:
% \begin{equation}
% p_{ij}
% =
% \frac{
% \exp(\mathrm{cos}(\hat{\mathbf{z}}_i, \mathbf{z}_j) / \tau)
% }{
% \sum_{k \in \mathcal{B}}
% \exp(\mathrm{cos}(\hat{\mathbf{z}}_i, \mathbf{z}_k) / \tau)
% }.
% \end{equation}
% The visual pretraining loss is
% \begin{equation}
% \mathcal{L}_{\mathrm{VP}}
% =
% -\frac{1}{|\mathcal{B}|}
% \sum_{i\in\mathcal{B}}
% \log p_{ii}.
% \label{eq:VP_loss}
% \end{equation}
% Here, $\hat{\mathbf{z}}_i$ is produced from the LLM hidden state at a visual position, 
% while $\mathbf{z}_i$ is the corresponding next visual latent. 
% Other target latents in $\mathcal{B}$ serve as in-batch negatives. 
% This objective encourages the model to predict the correct next visual latent while distinguishing it from other visual latents. 
% The expanded InfoNCE~\citep{oord2018representation} formulation and the construction of valid predicted--target pairs are provided in Appendix~\Cref{sec:appendix_pipeline}.

\paragraph{Joint text and visual pretraining}
The final training objective combines standard text next-token prediction with next visual latent prediction:
\begin{equation}
\mathcal{L}
=
\lambda_{\mathrm{text}}\mathcal{L}_{\mathrm{CE}}
+
\lambda_{\mathrm{vis}}\mathcal{L}_{\mathrm{VP}},
\label{eq:joint_objective}
\end{equation}
where $\mathcal{L}_{\mathrm{CE}}$ is the autoregressive cross-entropy loss on text tokens and $\mathcal{L}_{\mathrm{VP}}$ is defined in equation~\eqref{eq:VP_loss}. Text and visual examples are interleaved during training according to a fixed mixing ratio. We update the LLM, the visual input projection and the prediction head, while keeping the visual encoder frozen. Multiple foreground sequences are packed into fixed-length contexts, with sequence boundaries tracked to prevent cross-sample attention. Additional details on the training corpus, architecture, optimization setup and evaluation protocol are provided in Supplementary~\Cref{sec:appendix_details}.

\paragraph{Training setup}
% Continued pretraining is implemented based on XTuner framework~\citep{2023xtuner}.
% We conduct continued pretraining with XTuner~\citep{2023xtuner} under a distributed data-parallel setup.
We implement continued pretraining using the XTuner framework~\citep{2023xtuner}.
During visual pretraining, we optimize the LLM and the lightweight prediction projector that maps LLM hidden states to the frozen vision-representation space, while keeping the vision tower frozen.
% For the text-pretraining (TP) setting, we train on open-source text corpora with a total budget of 180B tokens.
% For the visual-pretraining (VP) setting, we train on 120B tokens constructed from sparse visual representations of document pages.
For both TP and VP, we keep the non-PDF text corpus, starting checkpoint, optimization recipe, and SFT stage fixed. 
The only controlled difference is the representation of the additional scientific-PDF corpus. 
In TP, the PDF pages are converted into MinerU2.5-parsed text, yielding approximately 80B text tokens. 
In VP, the same PDF pages are rendered as images and filtered into sparse foreground visual sequences, yielding approximately 20B visual tokens at the main resolution. 
Thus, the total CPT token budgets differ (180B for TP and 120B for VP) 
because the same matched PDF corpus is represented more compactly in the visual stream, 
not because VP uses a different document source. 
All comparisons therefore match the underlying PDF documents while allowing the token count to reflect the chosen representation.
The resulting checkpoints are then trained with the same SFT recipe and evaluated using the same benchmark protocols.

\paragraph{Evaluation protocol}
As shown in~\Cref{tab:main_result_text,tab:alignment_multimodal}, we evaluate \methodname and the baselines in a zero-shot setting after SFT initialized from CPT. 
The evaluations in~\Cref{tab:main_result_text} use CoT prompting with the instruction ``think step by step'', whereas those in~\Cref{tab:alignment_multimodal} use a direct-answer template without explicit reasoning guidance. 
Unless otherwise specified, \Cref{tab:main_result_text} reports average pass@8 for GPQA, the average score over 32 runs for AIME-25, and pass@1 for MMLU-Pro and HLE. 
All benchmarks in~\Cref{tab:alignment_multimodal} are reported with pass@1.
By default, we extract answers via rule-based matching and grade them automatically. When extraction fails, we use GPT-4o~\citep{hurst2024gpt} for judgment.

\clearpage
\bibliographystyle{iclr2025_conference}%
\bibliography{main}

\clearpage
\newpage
\clearpage
\newpage
\begin{appendices}
\renewcommand{\appendixname}{}
\makeatletter
\gdef\@seccntformat#1{\csname the#1\endcsname\quad}
\makeatother
\begin{strip}
  \begin{center}
  {\Large\bfseries \papertitle\par}
  \vspace{0.5em}
  {\Large\bfseries Supplementary Material\par}
  \vspace{1.0em}
\end{center}
\end{strip}

% \paragraph{Outline.}
\paragraph{Supplementary overview.}
\label{sec:appendix_overview}
We provide additional analyses and implementation details that complement the main text.

\begin{itemize}
  \item \textbf{Section~\ref{sec:appendix_pipeline} (Detailed Visual-Pretraining Pipeline).}
  We provide the full mathematical formulation of the visual pretraining pipeline, including frozen visual feature extraction, foreground-token filtering, causal visual prediction, 
  contrastive next-visual-latent loss, and sequence packing.
  % contrastive next-patch loss, and sequence packing.

  \item \textbf{Section~\ref{sec:appendix_retrieval} (PPL-Based Image-to-Text Retrieval).}
  We provide an alternative generative evaluation of cross-modal alignment through perplexity-based image-to-text retrieval, demonstrating that visual pretraining substantially enhances retrieval accuracy.

  % \item \textbf{Section~\ref{sec:appendix_attn} (Visual Token Analysis).}
  % We probe what visual tokens encode via attention visualization, demonstrating that pretrained visual tokens capture content-level and cross-region semantic structure that supports language reasoning.

  \item \textbf{Section~\ref{sec:appendix_genhead} (Further Studies: Visual Pretraining with Generative Decoder).}
  We explore augmenting visual pretraining with an explicit generative decoder that provides pixel-level reconstruction supervision. We find that this variant incurs substantially higher training cost while achieving performance comparable to the decoder-free visual pretraining formulation.

  \item \textbf{Section~\ref{sec:appendix_details} (Implementation and Evaluation Details).}
  We describe the training corpus, model architecture, optimization setup, evaluation protocol, and cross-modal alignment analysis used in the main experiments.

  % \item \textbf{Section~\ref{sec:related} (Related Work).}
  % We discuss prior work on math-focused continued pretraining, optical language modelling, and multimodal LLMs, and position \methodname as a label-free visual pretraining paradigm for improving both language and multimodal reasoning from raw scientific documents.
\end{itemize}

\section{Detailed Visual-Pretraining Pipeline}
\label{sec:appendix_pipeline}

This section provides the mathematical details of the visual-pretraining pipeline used in \methodname. The main text describes the method at a high level; here we specify the construction of sparse document tokens, the causal visual-prediction objective, and the sequence-packing procedure used in training.

\paragraph{Frozen visual feature extraction.}
Given a rendered document page $\mathcal{I}\in\mathbb{R}^{H\times W\times 3}$, a frozen ViT encoder $\phi_{\mathrm{ViT}}$ with patch size $p$ first maps the page into a grid of patch features:
\begin{equation}
Y
=
\phi_{\mathrm{ViT}}(\mathcal{I})
=
(y_1,\ldots,y_{N_0}),
\qquad
y_i\in\mathbb{R}^{c_v}.
\label{eq:appendix_vit_features}
\end{equation}
A frozen spatial merger $M$ then groups neighbouring patch features and maps them into the visual feature space used as the prediction target:
\begin{equation}
\mathcal{Z}
=
M(Y)
=
(z_1,\ldots,z_N),
\qquad
z_i\in\mathbb{R}^{d_v}.
\label{eq:appendix_merged_features}
\end{equation}
The ViT encoder and the spatial merger are fixed throughout training. Thus, the visual stream learns to predict stable document features rather than reconstructing pixels.

\paragraph{Foreground-token filtering.}
Document pages contain large blank regions, such as margins and whitespace. To avoid spending context length on background patches, we compute a patch-level foreground mask before constructing the visual sequence. For each raw image patch, we compute its pixel variance $\sigma_i^2$ and average luminance $\ell_i$. A patch is treated as background when it has low variance and near-blank luminance:
\begin{equation}
b_i
=
\mathbf{1}
\left[
\sigma_i^2<\tau_{\sigma}
\;\land\;
\left(
\ell_i>\tau_{\ell}^{+}
\lor
\ell_i<\tau_{\ell}^{-}
\right)
\right],
\label{eq:appendix_background_mask}
\end{equation}
where $b_i=1$ indicates background. After spatial merging, the mask is max-pooled over each merged block, so a merged visual token is retained if any of its constituent patches contains foreground content:
\begin{equation}
m_j
=
\max_{i\in\mathcal{P}(j)}
(1-b_i),
\label{eq:appendix_merged_mask}
\end{equation}
where $\mathcal{P}(j)$ denotes the set of raw patches belonging to merged token $j$.

The retained visual features are ordered in raster scan:
\begin{equation}
\mathcal{U}
=
\operatorname{Raster}\{z_j:m_j=1\}
=
(u_1,\ldots,u_L),
\qquad
L\ll N .
\label{eq:appendix_foreground_sequence}
\end{equation}
This produces a sparse document sequence that preserves page order while removing most blank regions.

\paragraph{Projection into the LLM space.}
Each foreground visual feature is projected into the LLM hidden space by a learned linear map:
\begin{equation}
x_i
=
W_{\mathrm{in}}u_i,
\qquad
x_i\in\mathbb{R}^{d}.
\label{eq:appendix_visual_projection}
\end{equation}
Position indices are assigned according to the raster order after foreground filtering. This allows the LLM to process the page as an ordered visual sequence while avoiding position allocation to removed background regions.

\paragraph{Causal visual prediction.}
The projected sequence $(x_1,\ldots,x_L)$ is fed into the shared autoregressive LLM backbone with a causal attention mask over image positions. The hidden state at position $i$ is used to predict the next foreground feature:
\begin{equation}
h_i
=
\Phi_{\mathrm{LLM}}(x_{\leq i}),
\qquad
\hat{u}_{i+1}
=
\psi(h_i),
\label{eq:appendix_vp_forward}
\end{equation}
where $\Phi_{\mathrm{LLM}}$ is the LLM backbone and $\psi$ is a lightweight MLP prediction head. In our implementation,
\begin{equation}
\psi(h)
=
W_2\,\mathrm{GELU}(W_1h).
\label{eq:appendix_prediction_head}
\end{equation}
The target for $\hat{u}_{i+1}$ is the frozen visual feature $u_{i+1}$. Therefore, the task is a continuous analogue of language-model next-token prediction.

\paragraph{Contrastive next-visual-latent loss.}
For a batch of valid predicted--target pairs $\mathcal{B}$, we compute cosine-similarity logits:
\begin{equation}
s_{ij}
=
\frac{\cos(\hat{u}_{i+1},u_{j+1})}{\tau},
\label{eq:appendix_vp_logits}
\end{equation}
where $\tau$ is a temperature parameter. The matching probability is obtained by a softmax over target features in the batch:
\begin{equation}
p_{ij}
=
\operatorname{softmax}_{j\in\mathcal{B}}(s_{ij}).
\label{eq:appendix_vp_softmax}
\end{equation}
The visual pretraining loss is then
\begin{equation}
\mathcal{L}_{\mathrm{VP}}
=
-\frac{1}{|\mathcal{B}|}
\sum_{i\in\mathcal{B}}
\log p_{ii}.
\label{eq:appendix_vp_loss}
\end{equation}
This objective encourages each predicted feature to match the correct next foreground token while using other visual features in the batch as negatives.

\paragraph{Joint optimization.}
Visual pretraining is combined with standard text next-token prediction:
\begin{equation}
\mathcal{L}
=
\lambda_{\mathrm{text}}\mathcal{L}_{\mathrm{CE}}
+
\lambda_{\mathrm{vis}}\mathcal{L}_{\mathrm{VP}}.
\label{eq:appendix_joint_loss}
\end{equation}
Only the LLM parameters, the visual input projection $W_{\mathrm{in}}$, and the prediction head $\psi$ are updated. The ViT encoder and spatial merger remain frozen, providing a fixed visual target space.

\paragraph{Sequence packing and attention masking.}
To improve training efficiency, multiple sparse foreground sequences are packed into a fixed-length context. Let $q(a)$ denote the document sequence to which packed position $a$ belongs. We use a block-causal attention mask:
\begin{equation}
A_{ab}
=
\begin{cases}
0, & q(a)=q(b)\ \text{and}\ b\leq a,\\
-\infty, & \text{otherwise}.
\end{cases}
\label{eq:appendix_block_causal_mask}
\end{equation}
This prevents visual tokens from attending to future tokens or to tokens from other packed document pages. Prediction targets are also restricted within the same original foreground sequence, so the model never predicts across document boundaries.
\section{PPL-Based Image-to-Text Retrieval}\label{sec:appendix_retrieval}

We provide an alternative, generative evaluation of cross-modal alignment by casting it as image-to-text retrieval scored by conditional perplexity.

\paragraph{Evaluation setting.}
We sample 100 random document image--text pairs and, for each (image $v_i$, text $t_j$) combination, compute a matching score defined following \citet{lin2023revisiting} as the pointwise mutual information estimated via conditional perplexity, $\text{Score}(i,j) = \mathcal{L}(t_j \mid v_i) - \mathcal{L}(t_j)$, where $\mathcal{L}(t_j \mid v_i)$ is the cross-entropy loss of generating text~$t_j$ conditioned on visual representation~$v_i$ and $\mathcal{L}(t_j)$ is the unconditional text prior; a lower score indicates stronger alignment. We report Recall@$K$ and Mean Reciprocal Rank (MRR).

\paragraph{Findings.}
As shown in \Cref{tab:i2t-retrieval}, visual pretraining yields a substantial improvement: R@1 increases from 64.0\% to 99.0\% and MRR rises from 78.2 to 99.5. Analysis of the $100 \times 100$ score matrix reveals that in the original model, off-diagonal scores exhibit high variance ($\sigma = 0.168$), causing certain texts to act as ``hub'' attractors that receive spuriously low scores regardless of the input image. After visual pretraining, the off-diagonal variance drops to $\sigma = 0.043$, effectively eliminating the hub phenomenon and enabling near-perfect discrimination.

\begin{table}[h]
  \centering
  \caption{\textbf{Generative image-to-text retrieval with conditional perplexity.}
  Each image is matched against 100 candidate texts using the PMI-style score defined in~\Cref{sec:appendix_retrieval}; higher Recall@$K$ and MRR indicate better cross-modal compatibility.}
  \label{tab:i2t-retrieval}
  \setlength{\tabcolsep}{4.3pt}
  \renewcommand{\arraystretch}{1.15}
  \begin{tabular}{lccc}
    \toprule
    Metric & Text Pretraining & VP (Ours) & $\Delta$ \\
    \midrule
    R@1  & \heatb{64.0}{64.0}{99.0}  & \heatb{99.0}{64.0}{99.0}  & \gooddelta{+35.0} \\
    R@5  & \heatb{92.0}{92.0}{100.0} & \heatb{100.0}{92.0}{100.0} & \gooddelta{+8.0}  \\
    MRR  & \heatb{78.2}{78.2}{99.5}  & \heatb{99.5}{78.2}{99.5}  & \gooddelta{+21.3} \\
    \bottomrule
  \end{tabular}
\end{table}
\section{Visual Pretraining with Generative Decoder}\label{sec:appendix_genhead}

\begin{figure*}[t]
  \centering
  \includegraphics[width=\linewidth]{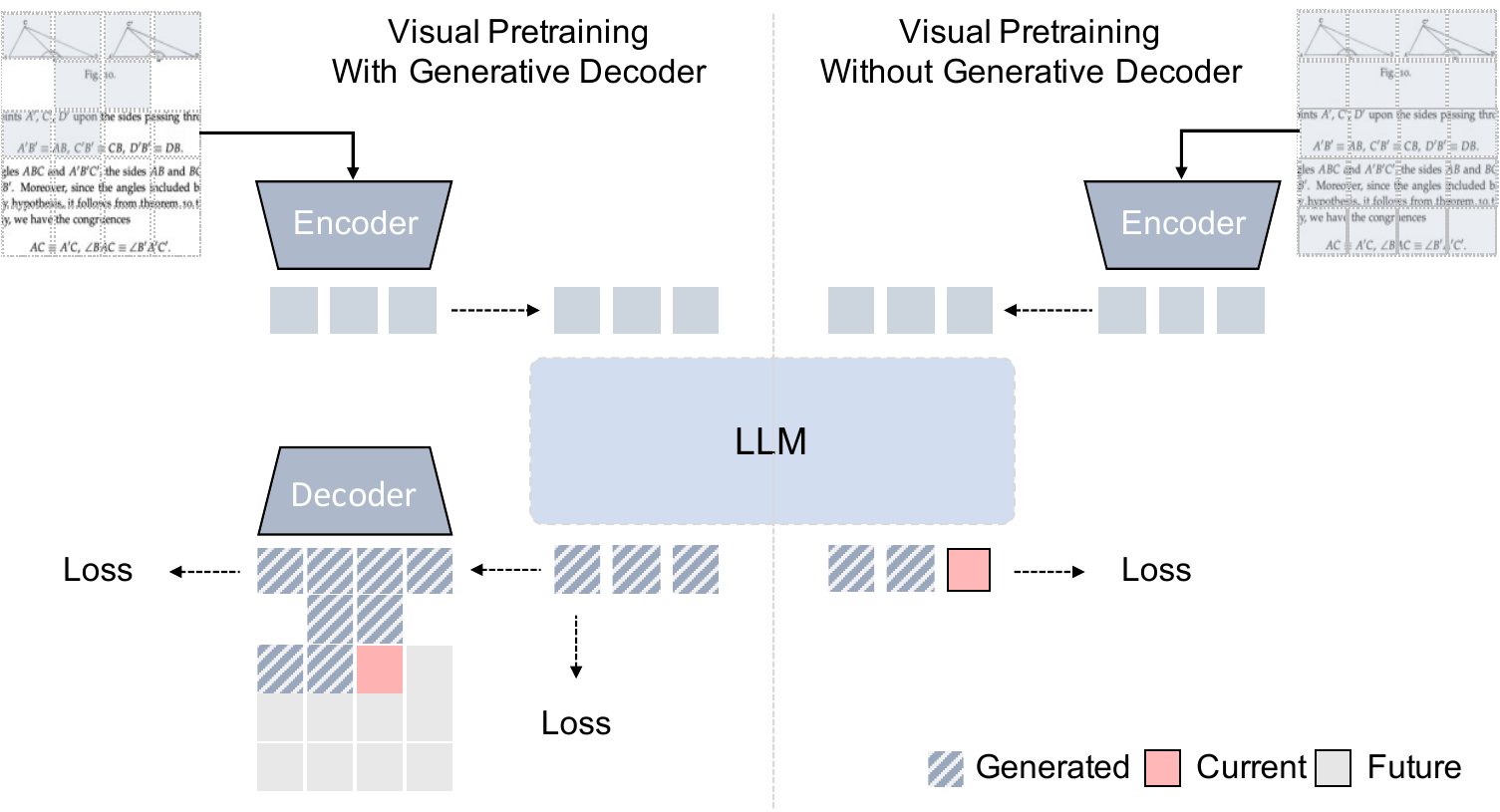}
  \caption{\textbf{Architecture comparison between the generative-decoder variant and decoder-free visual pretraining.}
  (Left) The generative-decoder variant (VP with Decoder) adds a MAR decoder and a frozen VAE decoder to reconstruct pages from LLM-refined latents, providing additional pixel-level supervision at the cost of extra parameters and computation.
  (Right) Our decoder-free formulation (VP w/o Decoder) predicts foreground patch latents autoregressively without pixel reconstruction.}
  \label{fig:comp_gen_head}
\end{figure*}

\noindent\textbf{Motivation.}
A natural extension of visual pretraining is to equip the LLM with an explicit \emph{generative decoder} that decodes visual latents back to pixels, thereby providing pixel-level reconstruction supervision \citep{zhao2026exploring}.
In our exploratory experiments, we instantiate this design by attaching a MAR-style autoregressive decoder~\citep{li2024autoregressive} followed by a frozen VAE decoder to the LLM's hidden states, yielding a full encode--decode pipeline as illustrated in \Cref{fig:comp_gen_head}.
The resulting pixel reconstruction loss $\mathcal{L}_{\text{MSE}}$ offers fine-grained, pixel-level supervision that is absent in the pure latent-prediction formulation used in the main text.

\noindent\textbf{Comparison with generative visual pretraining.}
Despite the extra supervision signal, the generative-decoder design incurs a substantial increase in training cost.
Specifically, (1) the MAR decoder introduces additional trainable parameters ($\sim$\,300M in our setup) and computation; (2) the diffusion-style latent loss $\mathcal{L}_{\text{diff}}$ and the pixel MSE loss $\mathcal{L}_{\text{MSE}}$ must both be computed and back-propagated through the decoder stack at every visual step; and (3) the overall training throughput drops by roughly $30$\%--$40$\% compared with the decoder-free variant.
In our experiments, the generative-decoder variant requires approximately $1.4\times$ the training time to reach the same convergence criterion as the decoder-free formulation.
These overheads become significant when scaling to large document corpora.

We further compare the generative-decoder variant (denoted \textbf{VP with Decoder}) against the decoder-free visual pretraining (denoted \textbf{VP w/o Decoder}) on the same matched document sources.
As summarized in \Cref{tab:genhead}, both visual pretraining variants outperform the Base model and the text-only pretraining baseline across all evaluated reasoning benchmarks.
Crucially, the pixel-level loss does \emph{not} translate into consistently stronger language-reasoning performance: the generative-decoder variant achieves comparable or only marginally better scores than the decoder-free formulation, suggesting that the extra pixel supervision yields diminishing returns for downstream reasoning.

\begin{table}[htbp]
\centering
\caption{\textbf{Decoder-free visual pretraining versus the generative-decoder variant.}
All methods use the same matched document sources and are evaluated after the same SFT protocol.
The decoder-free formulation attains comparable or better reasoning performance without the additional reconstruction decoder.}
\label{tab:genhead}
\setlength{\tabcolsep}{5pt}
\renewcommand{\arraystretch}{1.08}
\begin{tabular}{lccc}
\toprule
\textbf{Method} & \textbf{MMLU-Pro} & \textbf{GPQA} & \textbf{AIME} \\
\midrule
Base & \heatb{79.86}{79.86}{81.87} & \heatb{69.57}{69.57}{75.95} & \heatb{74.90}{74.48}{76.98} \\
Text pretraining & \heatb{81.32}{79.86}{81.87} & \heatb{74.94}{69.57}{75.95} & \heatb{74.48}{74.48}{76.98} \\
VP with Decoder & {\heatb{81.87}{79.86}{81.87}} & \heatb{75.44}{69.57}{75.95} & {\heatb{76.46}{74.48}{76.98}} \\
VP w/o Decoder & \heatb{81.69}{79.86}{81.87} & {\heatb{75.95}{69.57}{75.95}} & {\heatb{76.98}{74.48}{76.98}} \\
\bottomrule
\end{tabular}
\end{table}

\noindent\textbf{Conclusion.}
Taken together, these exploratory results suggest that pixel-level generation supervision offers limited marginal benefit for language reasoning, at a non-negligible computational cost.
Consequently, the main text adopts the \emph{decoder-free} visual pretraining formulation, which attains comparable downstream performance through a substantially simpler and more efficient pipeline.
\section{Additional Implementation and Evaluation Details}
\label{sec:appendix_details}

This section provides implementation and evaluation details that complement the main Methods. We first describe the training data, model architecture, optimization setup, and benchmark protocol. We then give the full construction of the cross-modal alignment analysis reported in~\Cref{tab:alignment_multimodal} and discussed in~\Cref{sec:results_alignment}.

\paragraph{Training corpus.}
\label{sec:appendix_corpus}

The training data consist of three components: a text corpus for standard continued pretraining, a scientific-PDF corpus for matched visual and text pretraining, and SFT data for instruction tuning. For the matched text-pretraining baseline, each PDF page is converted into text with MinerU2.5~\citep{wang2024mineru}. For \methodname, the same PDF pages are rendered as images and consumed as unlabeled visual supervision.
For the pretraining of Qwen, the actual preprocessing pipeline yields approximately 20B retained visual tokens for \methodname and approximately 80B parsed text tokens for TP from the same PDF pages. This matched construction controls the document source and changes only the representation of the scientific-PDF corpus.

The VP image records are stored as JSONL entries containing page-image paths and optional precomputed latent-feature paths. 
During loading, each page is converted to RGB, resized to a square at a fixed maximum resolution, and normalized before being passed through the model processor. Unless otherwise stated, pages are sampled without filtering by crop coordinates. Blank regions are removed by a foreground mask computed from patch variance and luminance, using variance threshold $0.02$, high-luminance threshold $0.95$, and low-luminance threshold $0.15$. A merged visual token is retained if any of its constituent raw patches is foreground.

\paragraph{Model architecture.}
\label{sec:appendix_architecture}

\methodname uses the same LLM backbone as the corresponding text-pretraining baseline. In the Qwen3.5 implementation, the visual pathway is initialized from the corresponding Qwen3.5/Qwen-VL checkpoint. The vision tower is a ViT-style encoder with 27 layers, hidden size 1152, 16 attention heads, patch size 16, and spatial merge size 2. The vision tower is frozen during VP. The foreground visual features are mapped into the LLM hidden space through a visual projector; in the main VP runs, this projector and the LLM are trainable, while the frozen vision features serve as the prediction targets. A two-layer MLP prediction head maps LLM hidden states back to the frozen visual-feature space for next visual latent prediction.

The main formulation of \methodname is decoder-free: it predicts frozen visual features and does not reconstruct pixels. This avoids the additional computational cost and optimization complexity of a pixel-level image decoder. We separately study a generative-decoder variant in~\Cref{sec:appendix_genhead}.

\paragraph{Training hyperparameters.}
\label{sec:appendix_training_hparams}

Text and visual examples are interleaved during continued pretraining according to a fixed mixing ratio. In our implementation, one VP batch is attached to every text-training step. The text branch is optimized with standard autoregressive cross-entropy, while the visual branch is optimized with the next visual latent contrastive loss defined in~\Cref{eq:VP_loss}. We use InfoNCE with temperature $\tau=0.07$ and set the VP loss weight to $0.1$ in the main Qwen3.5 runs. The LLM backbone, visual projector, and prediction head are updated jointly, while the vision tower remains frozen throughout training.

Multiple sparse foreground sequences are packed into fixed-length contexts, and sequence boundaries are tracked to prevent attention across different document pages. The detailed mathematical formulation of the visual pipeline, including foreground filtering, projection, causal masking, and sequence packing, is provided in~\Cref{sec:appendix_pipeline}.

For the Qwen3.5 runs, text sequences are hard-packed to 32,768 tokens and trained for 12,000 steps with global batch size 1024, AdamW learning rate $3\times10^{-5}$, weight decay $0.1$, a 1,000-step warmup, and linear decay to $10^{-6}$. The corresponding VP head uses AdamW with learning rate $4\times10^{-5}$ and the same optimizer betas and weight decay as the text optimizer. The visual VP context is capped at 8,192 retained foreground tokens for the main setting. 
% For the large-scale Qwen3.5 397B runs, text sequences are hard-packed to 256K tokens and trained for 30,000 steps on 512 distributed workers with sequence parallel size 4, global batch size 128, Muon learning rate $3\times10^{-5}$, weight decay $0.1$, a 1,000-step warmup, constant learning rate thereafter, FP8 tilewise scaling enabled, and a 32,768-token VP visual context.

\paragraph{Training cost.}
Across the main Qwen and Llama VP/TP continued-pretraining runs, each run typically required approximately 1.5--3 days of wall-clock time on a distributed cluster with 128 accelerators, depending on the backbone,
token budget, context length, and data modality.
For example, the main Qwen3.5-35B-A3B continued-pretraining run required approximately 36 hours, followed by approximately 10 hours for the SFT stage.
The exact accelerator model is not disclosed due to institutional constraints.

\paragraph{Evaluation protocol.}
\label{sec:appendix_eval_protocol}

All models are evaluated in a zero-shot setting after SFT initialized from CPT. For text-only reasoning benchmarks in~\Cref{tab:main_result_text}, we use CoT prompting with the instruction ``think step by step''. Unless otherwise specified, we report average pass@8 for GPQA, the average score over 32 runs for AIME-25, and pass@1 for MMLU-Pro and HLE.

For multimodal benchmarks in~\Cref{tab:alignment_multimodal}, we use a direct-answer template without explicit reasoning guidance and report pass@1. These evaluations test whether the same unlabeled visual-document pretraining signal that improves language reasoning also transfers to multimodal reasoning.

\paragraph{Cross-modal alignment analysis.}
\label{sec:appendix_alignment_details}

This paragraph details the cross-modal alignment evaluation reported in~\Cref{tab:alignment_multimodal} and qualitatively visualized in~\Cref{fig:main}.

\paragraph{Setting.}
We construct a set of $N{=}100$ matched document image--text pairs. 
Each pair is obtained by rendering a scientific PDF page as a RGB image and using the OCR-parsed text from the same page as its textual counterpart. The pairs are drawn from a held-out set of scientific PDFs collected from publicly accessible repositories.

For each pair, the image is processed by the visual pathway and the text by the tokenizer of the shared backbone. We collect hidden states from the last Transformer layer and mean-pool over foreground visual tokens and non-padding text tokens, obtaining one embedding per modality. This yields paired embeddings
\[
\{(v_i,t_i)\}_{i=1}^{N},
\]
where $v_i$ and $t_i$ denote the visual and textual representations of the same document page. The same procedure is applied before and after \methodname, enabling a direct comparison of representation alignment.

\paragraph{Metric definitions.}
We measure alignment from three complementary perspectives: global distance, structural geometry, and local neighbourhood consistency.

Global alignment is measured by centroid separation and pairwise cosine similarity. Centroid separation is defined as
\[
\|\bar{v}-\bar{t}\|_2,
\]
where $\bar{v}=\frac{1}{N}\sum_i v_i$ and $\bar{t}=\frac{1}{N}\sum_i t_i$ are the modality-wise mean embeddings. Pairwise cosine similarity is computed as
\[
\frac{1}{N}\sum_{i=1}^{N}\cos(v_i,t_i).
\]

Structural alignment is measured by linear Centered Kernel Alignment (CKA)~\citep{kornblith2019similarity}. Let $V,T\in\mathbb{R}^{N\times d}$ denote the centered embedding matrices of the visual and textual representations. We compute
\[
\mathrm{CKA}(V,T)
=
\frac{\|V^\top T\|_F^2}
{\|V^\top V\|_F\,\|T^\top T\|_F}.
\]
Linear CKA compares the relational geometry of the two embedding spaces and is invariant to orthogonal transformations and isotropic scaling.

Local alignment is measured by mutual $k$-nearest-neighbour overlap:
\[
A_k
=
\frac{1}{Nk}
\sum_{i=1}^{N}
\left|
\mathcal{N}^{V}_{k}(i)
\cap
\mathcal{N}^{T}_{k}(i)
\right|,
\]
where $\mathcal{N}^{V}_{k}(i)$ and $\mathcal{N}^{T}_{k}(i)$ are the $k$ nearest neighbours of sample $i$ in the visual and textual embedding spaces under cosine distance. We report $k\in\{1,5,10\}$.

\paragraph{Findings.}
The three groups of metrics provide complementary evidence that \methodname improves cross-modal alignment without image--text pair supervision. Globally, centroid separation drops by $60\%$ from $1.665$ to $0.661$, while pairwise cosine similarity rises from $0.631$ to $0.907$. This indicates that paired visual and textual embeddings are brought into a more compatible region of representation space.

Structurally, linear CKA increases from $0.657$ to $0.745$, showing that the improvement is not merely a shift in average distance but also a change in the relational geometry of the two modalities. Locally, mutual $k$-NN overlap improves at all tested values of $k$, with the largest gain at the most stringent setting: $k{=}1$ increases from $0.140$ to $0.310$, while $k{=}5$ increases from $0.288$ to $0.420$ and $k{=}10$ from $0.395$ to $0.496$.

This pattern argues against trivial representation collapse. Collapse would tend to inflate neighbourhood overlap uniformly, whereas \methodname produces the strongest improvement at the finest local scale. Together, the global, structural, and local metrics indicate that visual pretraining reshapes the shared representation space in a way that makes visual and textual document states more compatible.
\end{appendices}

\end{document}